\documentclass[11pt]{article}

\PassOptionsToPackage{usenames,dvipsnames}{xcolor}
\usepackage[final]{acl}

\usepackage{times}
\usepackage{latexsym}

\usepackage[T1]{fontenc}
\usepackage[utf8]{inputenc}

\usepackage{microtype}

\usepackage{inconsolata}

\usepackage{graphicx}
\usepackage{tikz}
\usetikzlibrary{arrows.meta,positioning}

\usepackage{booktabs}
\usepackage{amsfonts}
\usepackage{amssymb}
\usepackage{mathtools}
\usepackage{nicefrac}
\usepackage{enumitem}
\usepackage{subcaption}
\usepackage{comment}
\usepackage{array}
\usepackage{float}
\usepackage{wrapfig}
\usepackage{xspace}
\usepackage{bm}
\usepackage{suffix}
\usepackage{algorithm}
\usepackage[noend]{algpseudocode}
\usepackage{mdframed}
\usepackage[usenames,dvipsnames]{xcolor}
\usepackage[textwidth=1.5cm]{todonotes}
\usepackage{listings}
\usepackage[noabbrev,capitalize]{cleveref}  % must be loaded after hyperref (via acl.sty)

\newcommand{\rightcomment}[1]{\(\triangleright\) {\small \it #1}}
\newcommand{\eqcomment}[1]{\addtocounter{equation}{1}\tag*{\rightcomment{#1}\quad(\theequation)}}
\WithSuffix\newcommand\eqcomment*[1]{\tag*{\rightcomment{#1}}}

\renewcommand\algorithmicthen{:}
\algnewcommand{\IfThen}[2]{\State \algorithmicif\ #1\ \algorithmicthen\ #2}
\algnewcommand{\IfThenElse}[3]{\State \algorithmicif\ #1\ \algorithmicthen\ #2\ \algorithmicelse\ #3}
\algrenewcommand{\algorithmiccomment}[1]{\hfill \rightcomment{#1}}
\algnewcommand{\LineComment}[1]{\State \rightcomment{#1}}
\algnewcommand{\LinesComment}[1]{\State \rightcomment{\parbox[t]{\linewidth-\leftmargin-\widthof{\(\triangleright\) }}{#1}}\smallskip}
\algnewcommand\algorithmicinput{{\bfseries Input:}}
\algnewcommand\INPUT{\item[\algorithmicinput]}
\algnewcommand\algorithmicoutput{{\bfseries Output:}}
\algnewcommand\OUTPUT{\item[\algorithmicoutput]}
\makeatletter
\newcounter{algorithmicH}
\let\oldalgorithmic\algorithmic
\renewcommand{\algorithmic}{%
  \stepcounter{algorithmicH}
  \oldalgorithmic}
\renewcommand{\theHALG@line}{ALG@line.\thealgorithmicH.\arabic{ALG@line}}
\makeatother
\makeatletter
\newcommand{\algmargin}{\the\ALG@thistlm}
\makeatother
\algnewcommand{\Statepar}[1]{\State\parbox[t]{\dimexpr\linewidth-\algmargin}{\strut #1\strut}}

\newcommand{\para}[1]{\noindent \textbf{#1}}
\newcommand{\cutforspace}[1]{}

\definecolor{aigold}{RGB}{244,210,1}
\definecolor{aigreen}{RGB}{213,245,227}
\definecolor{humanpurple}{RGB}{235,222,240}
\definecolor{mypurple}{RGB}{147,112,219}
\definecolor{myorange}{RGB}{255,165,0}
\definecolor{commentgray}{RGB}{86,101,115}
\definecolor{mygray}{RGB}{169,169,169}
\definecolor{aired}{RGB}{255,180,181}
\definecolor{pastelgreen}{RGB}{50,205,50}

\crefname{equation}{equation}{equations}
\crefname{footnote}{footnote}{footnotes}
\crefname{listing}{Example}{Examples}
\crefname{assumption}{assumption}{assumptions}
\crefname{line}{line}{lines}
\crefname{section}{\S}{\S\S}

\newcommand{\increase}{\textcolor{pastelgreen}{\bm{$\uparrow$}}}
\newcommand{\decrease}{\textcolor{red}{\bm{$\downarrow$}}}

\title{A Single Rewrite Suffices: Empirical Lessons from Production Skill Description Optimization}

\author{Yangqiaoyu Zhou, Mohammad Alqudah, Kwei-Herng Lai, Aaron Halfaker \\ {\bf Yingqi Xiong, Yaar Harari} \\
        Microsoft \\ \texttt{\{yangqzhou, yaarharari\}@microsoft.com}}

\begin{document}
\maketitle
\begin{abstract}

Enterprise AI agents route user queries to specialized skills by matching queries against natural language skill descriptions. When two skills share overlapping descriptions, the routing LLM misroutes queries, a failure we term skill collision. As agents scale to dozens of skills, manually tuning descriptions to maintain routing accuracy becomes a significant engineering bottleneck.
We deploy an automated description optimization pipeline on a production enterprise group chat agent (9 skills, 372 regression cases). The pipeline produces descriptions averaging 79.2\% F1, matching manually tuned descriptions at 79.4\% F1 (average per-skill difference -0.20\%, within the $\pm0.78\%$ multi-seed noise floor), while reducing per-skill engineering effort from 120 minutes to 3.8 minutes (32$\times$ speedup).
We then examine which pipeline components actually drive this match. 
Systematic ablation on both the production system and ToolBench ($\sim$16k tools) reveals that a single LLM rewrite using any available false-positive and false-negative cases captures most of the available improvement. Other design choices we tested (iteration budget, feedback signal composition, dual editing of confused pairs, and training set size) each affect final F1 by less than 0.5\%.
Description optimization addresses skill collisions caused by overlapping descriptions but cannot resolve cases where two skills' intended scopes genuinely overlap. We identify a diagnostic (a large train–validation F1 gap) that flags the latter cases for architectural rather than text-level intervention.
\end{abstract}

\section{Introduction}
\label{sec:introduction}

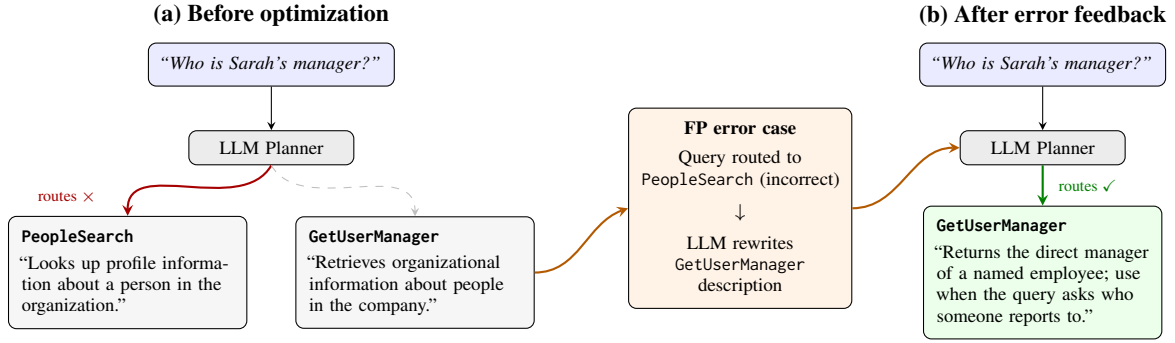
\begin{figure*}[t]
\centering
\begin{tikzpicture}[
  font=\scriptsize,
  >=stealth,
  qbox/.style={draw, rounded corners=3pt, fill=blue!8, text width=2.9cm, align=center, inner sep=5pt},
  pbox/.style={draw, rounded corners=3pt, fill=gray!15, text width=1.9cm, align=center, inner sep=4pt},
  skbox/.style={draw, rounded corners=3pt, fill=gray!7, text width=2.8cm, align=left, inner sep=5pt},
  skboxg/.style={draw, rounded corners=3pt, fill=green!8, text width=2.8cm, align=left, inner sep=5pt},
  fbbox/.style={draw, rounded corners=3pt, fill=orange!10, text width=2.6cm, align=center, inner sep=5pt},
]
%% nodes
\node[font=\small\bfseries] at (2.3, 0.65) {(a)~Before optimization};
\node[qbox] (q1) at (2.3, 0) {\textit{``Who is Sarah's manager?''}};
\node[pbox] (p1) at (2.3, -1.1) {LLM Planner};
\node[skbox] (skL) at (0.4, -2.75) {\textbf{\texttt{PeopleSearch}}\\[2pt]``Looks up profile information about a person in the organization.''};
\node[skbox] (skR) at (4.2, -2.75) {\textbf{\texttt{GetUserManager}}\\[2pt]``Retrieves organizational information about people in the company.''};
\node[fbbox] (fb) at (8.5, -1.9) {%
  \textbf{FP error case}\\[3pt]
  Query routed to \texttt{PeopleSearch} (incorrect)\\[4pt]
  $\downarrow$\\[4pt]
  LLM rewrites \texttt{GetUserManager} description};
\node[font=\small\bfseries] at (12.5, 0.65) {(b)~After error feedback};
\node[qbox] (q2) at (12.5, 0) {\textit{``Who is Sarah's manager?''}};
\node[pbox] (p2) at (12.5, -1.1) {LLM Planner};
\node[skboxg] (skR2) at (12.5, -2.75) {\textbf{\texttt{GetUserManager}}\\[2pt]``Returns the direct manager of a named employee; use when the query asks who someone reports to.''};
%% arrows
\draw[->] (q1) -- (p1);
\draw[->, red!65!black, thick] (p1.south) to[out=240, in=90] (skL.north);
\node[font=\tiny, text=red!65!black] at (-0.4, -1.75) {routes~$\times$};
\draw[->, gray!50, dashed] (p1.south) to[out=300, in=90] (skR.north);
\draw[->, orange!70!black, thick] (skR.east) to[out=0, in=200] (fb.west);
\draw[->, orange!70!black, thick] (fb.east) to[out=0, in=180] (p2.west);
\draw[->] (q2) -- (p2);
\draw[->, green!50!black, thick] (p2.south) -- (skR2.north)
  node[midway, right=2pt, font=\tiny, text=green!50!black] {routes~\checkmark};
\end{tikzpicture}
\caption{Skill collision and automated resolution via error feedback.
\textbf{(a)}~Before optimization: two skills share nearly identical descriptions; the LLM planner cannot distinguish them and misroutes the query to \texttt{PeopleSearch} (incorrect, $\times$).
\textbf{(b)}~After error-feedback-guided rewriting: the false-positive case is fed back to an LLM, which produces a discriminative description for \texttt{GetUserManager}; the planner now routes correctly (\checkmark).}
\label{fig:collision}
\end{figure*}

Modern enterprise AI assistants are built as orchestrated agent systems where an LLM-based planner routes user queries to specialized skills~\citep{yao2023react,wu2023autogen}.
Routing decisions are driven by comparing the query against each skill's natural language description.
When descriptions are insufficiently discriminative, the planner misroutes queries, causing what we term \emph{skill collision}: semantically overlapping skills compete for the same query population (Figure~\ref{fig:collision}).
 
Skill collision is most acute during onboarding.
Adding a new skill to a deployed agent forces its description to be precisely positioned relative to all incumbent skills.
In practice, this requires iterative manual tuning: a developer writes a description, deploys the skill, observes routing errors from real or synthetic traffic, and edits accordingly.
This loop is slow, requires expertise in both the skill's functionality and the planner's decision boundaries, and does not scale as agent capabilities grow.
 
We deploy an automated description optimization pipeline that replaces this manual loop. The pipeline initializes a candidate description with an LLM, then iteratively refines it using false-positive and false-negative cases from a labeled training set. On a production enterprise group chat agent (9 skills, 372 regression cases), the automated descriptions match manually tuned ones in routing quality (79.2\% vs. 79.4\% average F1, per-skill differences within the $\pm$0.78\% multi-seed noise floor) while reducing per-skill engineering effort from 120 minutes to 3.8 minutes (32$\times$ speedup).

Given that the automated pipeline reaches manual quality, we ask which of its components actually matter. As originally implemented, the pipeline combined several mechanisms motivated by intuition: contrastive feedback presenting FP, FN, and TP cases simultaneously; iterative refinement up to a fixed budget; dual editing of the most-confused rival skill; tuned training set sizes. Through systematic ablation on the production system and at scale on ToolBench~\citep{qin2023toolllm} ($\sim$16k tools), spanning closed-world (per-query candidate pool) and open-world (retrieval over the full corpus) routing regimes, we find that most of these design choices change final routing F1 by less than 0.5\%. 
A single LLM rewrite using any available FP/FN cases captures the bulk of available improvement: on production, single-shot achieves 79.2\% F1 (per-skill -0.12\% from human-written descriptions, matching iterative refinement within noise); on ToolBench open-world, single-shot achieves a +4.45\% F1 gain, within 0.2\% of iterative refinement.
 
Not all routing failures are description failures. Two further findings characterize where description optimization applies, and where it does not. First, closed-world (fixed candidate pool) and open-world (retrieval-based) routing require separate optimization: descriptions tuned in one regime transfer poorly to the other, and LLM initialization helps in one setting but hurts the other. Second, when two skills' intended scopes genuinely overlap (as opposed to merely sharing under-discriminative wording), no description rewrite can resolve the collision; these skills exhibit large training-validation F1 gaps regardless of optimization signal quality. We identify this gap as a diagnostic signal that flags skills for architectural intervention (scope restructuring, intent-specific routing rules) rather than continued description refinement.

Taken together, these findings yield a compact operational picture for skill onboarding: optimize with a single-shot LLM rewrite, match the optimization regime to the deployment regime, and use the train-validation F1 gap to triage which skills need architectural intervention instead of more description tuning. We make three contributions: (1) a deployed production system that replaces manual description tuning at enterprise scale, validated as non-inferior to manually tuned baselines with a 32$\times$ engineering speedup; (2) systematic ablation across production and ToolBench showing that pipeline complexity—feedback type, iteration budget, dual editing, training size—has negligible impact, and a single LLM rewrite suffices; (3) characterization of where description optimization applies, including a diagnostic signal for skills requiring architectural intervention and the separation of closed-world and open-world routing into distinct optimization regimes.

\section{Related Work}
\label{sec:related_work}

\para{Tool selection for LLM agents.}
LLM tool selection encompasses routing models and planners over large API collections~\citep{qin2023toolllm,shen2023hugginggpt,li2023apibank}, generating accurate calls from documentation~\citep{patil2023gorilla}, and retrieving candidate tools for orchestrators~\citep{liu2025toolscopeenhancingllmagent,jia2025autotoolefficienttoolselection,lumer2025tooltoagentretrievalbridgingtools}. These upstream retrieval methods complement our work; we optimize the descriptions used within the established candidate pools.

\para{Tool description optimization.}
Recent methods use LLMs and execution traces to rewrite tool documentation, aiming to improve downstream task completion~\citep{yuan2024easytoolenhancingllmbasedagents,quexploration,fang2025play2promptzeroshottoolinstruction,guo2026learningrewritetooldescriptions}. 
Instead of targeting downstream execution success via execution traces, we target upstream routing decisions using objective false positive and negative cases to diagnose skill collisions. Furthermore, we rigorously evaluate against manually tuned descriptions in a production deployment and demonstrate that a single LLM rewrite captures the vast majority of available improvements.

\para{Prompt optimization \& self-refinement}
Numerous frameworks optimize overarching task prompts or refine outputs using textual gradients, demonstrations, or LLM self-criticism \citep{khattab2024dspy,opsahlong2024optimizinginstructionsdemonstrationsmultistage,yang2024opro,zhou2023ape,Yuksekgonuletal2025,pryzant-etal-2023-automatic,madaan2023self,shinn2023reflexion,agrawal2026gepareflectivepromptevolution}. Instead of optimizing aggregate task prompts, we focus exclusively on per-skill descriptions evaluated via discrete routing errors. Furthermore, while many of these methods rely on multi-step evolutionary search or iterative closed-loop refinement, we demonstrate that a single-shot rewrite using objective routing feedback captures the vast majority of available improvements for tool description optimization.

\section{Method}
\label{sec:method}

The skill onboarding pipeline takes as input a skill name and outputs an optimized routing description for use by the LLM planner.
The pipeline has two stages: LLM initialization and error-feedback refinement.
We additionally describe a single-shot variant which section~\ref{subsec:simplification} shows is sufficient to match the full iterative pipeline.
 
\para{Stage 1: LLM initialization.}
We prompt an LLM with the skill name to generate a candidate routing description.
The initialization prompt requests a description of what the skill does. The initialized description serves as both the starting point for refinement and a standalone baseline for evaluation.

\para{Stage 2: Error-feedback refinement.}
The initialized description is evaluated on a labeled training set of queries with ground-truth skill labels.
We collect false positives (FP: queries routed to the skill that should not be), false negatives (FN: queries the skill misses), and true positives (TP: correctly routed queries).
At each iteration, we prompt the LLM with the current description, up to 5 FP and 5 FN cases (a practical token-budget choice; \S\ref{subsec:simplification} confirms performance is insensitive to this limit), and a TP set matched to the number of negative cases (to balance the prompt context across positive and negative examples).
The LLM is asked to identify routing failure patterns and revise the description.
The refined description replaces the current one, and the process repeats up to a fixed iteration budget or until per-skill training F1 exceeds a $90\%$ threshold (in practice, the iteration budget is the effective stopping criterion; see \S\ref{subsec:simplification}).
At each iteration $t$, the routing evaluation produces error cases $\mathcal{E}_t = (\text{FP}_t, \text{FN}_t, \text{TP}_t)$; the description with the highest training F1 across iterations is selected.
Algorithm~\ref{alg:refinement} summarizes the loop.
 
\begin{algorithm}[t]
\small
\caption{Error-Feedback Refinement}
\label{alg:refinement}
\begin{algorithmic}[1]
\INPUT skill $s$, training queries $Q$, budget $T$, threshold $\tau$
\OUTPUT optimized description $\hat{d}$
\State $d_0 \leftarrow \textsc{Initialize}(s)$ \Comment{Stage 1}
\State $\hat{d} \leftarrow d_0$, $\hat{f} \leftarrow 0$
\For{$t = 1, \ldots, T$}
    \State $f_t, \mathcal{E}_t \leftarrow \textsc{Evaluate}(d_{t-1}, Q)$
    \IfThen{$f_t > \hat{f}$}{$\hat{d} \leftarrow d_{t-1}$, $\hat{f} \leftarrow f_t$}
    \IfThen{$f_t \geq \tau$}{\textbf{break}}
    \State $d_t \leftarrow \textsc{LLMRewrite}(d_{t-1}, \textsc{Sample}(\mathcal{E}_t))$
\EndFor
\State \Return $\hat{d}$
\end{algorithmic}
\end{algorithm}

\para{Single-shot variant.}
We additionally evaluate a single-shot configuration where the LLM is given the initialized description and all available FP and FN cases in a single prompt, producing one revised description without further iteration. This simulates the approach a developer would take with a complete training set, and §5.2 shows it is sufficient to match iterative refinement.

\para{Open-world adaptation.}
In the retrieval setting, all tools are indexed by description and candidates are retrieved per query via hybrid sparse-dense retrieval (BM25~\citep{Robertson2009ThePR} combined with \texttt{text-embedding-ada-002}~\citep{openai2022ada} cosine similarity, with per-query min-max normalization).
The refinement loop runs identically, but FP cases now correspond to tools retrieved into the top-20 candidate pool that should not be invoked, providing a targeted disambiguation signal that reflects retrieval errors.
Only tools that appear in at least one FP query's retrieved pool are eligible for refinement.

\begin{table*}[t]
\centering
\small
\caption{
Per-skill skill-selection F1 on the production agent. Each row corresponds to replacing a single skill's description with an automatically generated variant, while keeping all other skill descriptions fixed at their human-written versions. HUMAN: all skills use the currently deployed, manually tuned descriptions (hence identical F1 across rows); INIT: the target skill's description is LLM-initialized with no error feedback; SS: single-shot LLM rewrite given all training FP/FN cases; Iter: iterative refinement loop (max 10 iterations). Average $\Delta$(HUMAN$\to$SS) of $-0.12\%$ and $\Delta$(HUMAN$\to$Iter) of $-0.20\%$ are both well within the multi-seed per-skill noise floor of $\pm 0.78\%$ (Appendix~\ref{sec:appendix-multiseed}). Skills sorted by $\Delta$(HUMAN$\to$Iter) in descending order.
}
\label{tab:prod_human}
\begin{tabular}{lrrrrrr}
\toprule
Skill & HUMAN & INIT & SS & Iter & $\Delta$(H$\to$SS) & $\Delta$(H$\to$Iter) \\
\midrule
IntKnowledge    & 79.4 & 77.1 & 80.1 & 81.0 & $+0.7$\increase                   & $+1.6$\increase \\
PeopleSearch    & 79.4 & 79.2 & 78.6 & 79.9 & $-0.8$\decrease                   & $+0.6$\increase \\
Email           & 79.4 & 78.6 & 80.4 & 79.8 & $+1.1$\increase                   & $+0.4$\increase \\
UserManager     & 79.4 & 79.4 & 78.8 & 79.7 & $-0.6$\decrease                   & $+0.4$\increase \\
WorkbackPlan    & 79.4 & 78.3 & 80.8 & 79.7 & $+1.5$\increase                   & $+0.3$\increase \\
DirectReports   & 79.4 & 79.5 & 79.1 & 78.6 & $-0.2$\decrease                   & $-0.7$\decrease \\
WebSearch       & 79.4 & 78.4 & 81.0 & 78.1 & $+1.6$\increase                   & $-1.3$\decrease \\
Calendar        & 79.4 & 77.6 & 76.9 & 77.9 & $-2.5$\decrease                   & $-1.5$\decrease \\
StatusReport    & 79.4 & 77.6 & 77.5 & 77.7 & $-1.9$\decrease                   & $-1.7$\decrease \\
\midrule
\textbf{Average} & \textbf{79.4} & 78.4 & 79.2 & 79.2 & $\mathbf{-0.12}$ & $\mathbf{-0.20}$ \\
\bottomrule
\end{tabular}
\end{table*}

\section{Experimental Setup}
\label{sec:experiments}

\para{Production setting.}
The production system is an enterprise group chat agent whose LLM-based planner routes queries to one of 9 skills spanning people search, web search, calendar scheduling, internal knowledge retrieval, organizational hierarchy lookup, email, and document generation.
We use 372 synthetic test cases with ground-truth skill labels, created by product and engineering experts.
Each query may target one or more skills; all targeted skills are labeled as positives.
Per-skill training queries are synthesized separately from the 372-case test set; positive examples range from 10 to 119 per skill depending on skill scope, with negative examples sampled from queries targeting other skills to ensure label balance. 
We compare four conditions: \textbf{HUMAN} (currently deployed manually tuned descriptions, our primary comparison baseline), \textbf{OLD} (skill disabled, system-level pre-deployment baseline), \textbf{INIT} (LLM-initialized description, no error feedback), and \textbf{Iter} (optimized description after refinement).
Training set is the full set of available examples per skill (referred to as \textbf{trainmax}); a smaller \textbf{train20} variant is reported in Appendix~\ref{sec:appendix-train20}.

\para{ToolBench setting.}
ToolBench~\citep{qin2023toolllm} covers $\sim$16k RESTful API tools across I1, I2, I3 splits.
I2 is the primary split for skill collision evaluation, because it requests multiple tools from the same API category to fulfill the query.
ToolBench descriptions are developer-facing API stubs (e.g., \emph{``Search Book by its name'', ``This endpoint will return back all news about Climate Change from all over the world''}) rather than routing instructions, which makes description quality directly measurable but also inflates the absolute gain numbers relative to systems with already-curated descriptions.
We frame the task as single-turn tool routing: given a query and a candidate pool, predict the correct tool, with no API execution.
We treat the first non-terminal tool call from the official ToolBench trajectories as ground truth; the original trajectories are LLM-generated and we did not manually verify each label.
We study two candidate pool regimes.
\textbf{Closed-world} uses the per-query pre-defined \texttt{available\_tools} list (3 to 15 tools) from the ToolBench dataset as the full candidate pool.
\textbf{Open-world} retrieves the top-20 candidates per query from the full $\sim$16k corpus via hybrid retrieval.
For each split, we select the top-100 tools by positive sample count (requiring at least 25 positive and 25 negative queries); these subsets are fixed across all ablations.
Per tool, we allocate 20 positive and 20 negative queries as the training set and hold out a balanced validation set of up to 50 positive and 50 negative queries, identical across all training-size ablation runs so that comparisons reflect only the change in training signal.
We track three description states: \textbf{ORIG} (iteration 0, original placeholder), \textbf{INIT} (iteration 1, LLM-generated with no error feedback), and \textbf{Iter} (best description across up to 10 refinement iterations).

\para{Ablation dimensions.}
On ToolBench we ablate: error signal type (FP+FN+TP / FP-only / FN-only / FP+FN without TP), iteration budget ($\{5, 10, 30\}$), per-iteration sampling (5 cases per category vs.\ all available), training size ($\{5, 10, 20\}$ examples per class), dual editing (simultaneously refining the rival tool's description), and three BM25 to dense retrieval weight combinations $(0.2, 0.8)$, $(0.5, 0.5)$, $(0.8, 0.2)$.
Default settings unless specified are train$=$20, max iterations$=$10, BM25$=$0.2, EMB$=$0.8.

\begin{table*}[t]
\centering
\small
\caption{Five design choices originally motivating the pipeline complexity, evaluated on ToolBench I2 open-world (BM25=0.2, EMB=0.8, max-10 iterations, train=20 unless ablated). $\Delta$ is the gain from orig to best across variants of each design choice. $n$ is the tools eligible for optimization (those with at least one retrieval false positive in the candidate pool); $n$ varies across rows because eligibility depends on the configuration. Detailed per-variant numbers in Appendix~\ref{sec:appendix-ablations}.}
\label{tab:robustness}
\begin{tabular}{llrr}
\toprule
Design choice & Variants & $\Delta$ range & $n$ \\
\midrule
Feedback signal     & FP+FN+TP / FP-only / FN-only / no TP   & $4.51$ - $4.60$ & 48 \\
Iteration vs.\ Single-Shot   & SS / max-10 sampled / max-10 full      & $4.45$ - $4.85$ & 47 \\
Dual editing        & on / off                                & $4.51$ - $4.54$    & 48 \\
Sampling            & 5 cases / all cases                     & $4.51$ - $4.85$ & 47 \\
Training size       & 5 / 10 / 20 per tool                   & $4.51$ - $4.92$ & 31--48 \\
\bottomrule
\end{tabular}
\end{table*}

\section{Results}
\label{sec:results}

\subsection{Automated optimization matches manual tuning at production scale}
\label{subsec:noninferior}

Table~\ref{tab:prod_human} reports per-skill HUMAN, INIT, single-shot (SS), and iterative (Iter) F1 on the production agent.
The average $\Delta$(HUMAN$\to$Iter) across the 9 skills is $-0.20\%$ and $\Delta$(HUMAN$\to$SS) is $-0.12\%$.
To characterize run-to-run variance of the live regression API, we ran 3 independent seeds of the trainmax pipeline; the average per-skill standard deviation of $\Delta$F1 is $\pm 0.78\%$, with 8 of 9 skills below $\pm 1\%$ std (Appendix~\ref{sec:appendix-multiseed}).
Both averages are well below this per-skill noise floor; the largest per-skill differences are $\pm 1.7\%$ for Iter and $\pm 2.5\%$ for SS (Table~\ref{tab:prod_human}).
Per-skill outcomes split nearly evenly under both methods: 5 skills favor Iter and 4 favor manual tuning; SS splits 4 vs.\ 5.
We interpret this as both single-shot and iterative automated optimization being non-inferior to manual tuning at production scale, but not exceeding it.

The value of automated onboarding is therefore operational.
An applied scientist manually tuned descriptions for one skill using the standard process (write, regression test, observe failures, edit), recorded wall-clock time, and compared against the automated pipeline. Manual tuning required 120 minutes for this skill, whereas the automated pipeline produced comparable descriptions in 3.8 minutes, which is a 32× speedup.

Notably, the single-shot variant achieves this match without iteration or any of the additional pipeline mechanisms. The next section examines which components of the pipeline actually contribute to this result.

\subsection{A single LLM rewrite captures most of the improvement}
\label{subsec:simplification}

Table~\ref{tab:robustness} summarizes ablations of five design choices originally motivating the elaborate pipeline: feedback signal composition, iteration budget, dual editing of confused tool pairs, per-iteration case sampling limit, and training set size.
All five vary final F1 within $0.5\%$ on ToolBench I2 open-world.
Specifically: feedback signal type (FP+FN+TP vs.\ FP-only vs.\ FN-only vs.\ FP+FN without TP) yields gains within $0.1\%$ of each other; dual editing provides negligible benefit; sampling 5 cases per category per iteration achieves similar gain as passing all available cases; training size effects (5 vs.\ 10 vs.\ 20 examples per tool) are within $0.5\%$, with train=10 marginally best.
The implication is that the per-example error feedback signal is the dominant driver of optimization quality. How that signal is packaged into the LLM rewrite prompt is has less effect.
Detailed per-variant numbers are in Appendix~\ref{sec:appendix-ablations}.

Figure~\ref{fig:iteration_budget} shows the iteration budget sweep.
On ToolBench open-world, gains increase monotonically with iteration budget: $+4.24\%$ at max-5, $+4.65\%$ at max-10, $+5.07\%$ at max-30.
Approximately $90\%$ of tools hit the iteration cap without reaching the $90\%$ stopping criterion at max-10, indicating optimization is still progressing when truncated.
However, single-shot rewriting (one LLM call given all available training FP and FN cases) achieves $+4.45\%$, matching max-10 within $0.2\%$ and exceeding max-5.
Iterative refinement with sampled feedback adds little when single-shot has access to the full training signal.
The production pattern reported in Section~\ref{subsec:noninferior} confirms this finding at production skill.

For practitioners trading compute against performance, single-shot is the dominant choice across both settings. Long iteration budgets capture marginal additional gains in open-world ToolBench but the additional cost is rarely justified.

\begin{figure}[t]
\centering
\includegraphics[width=\columnwidth]{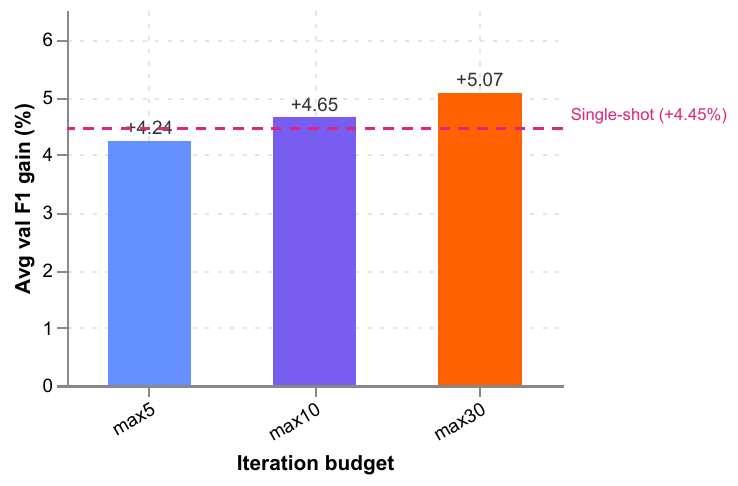}
\caption{Average F1 gain on validation across different max iteration budget on ToolBench I2 open-world (BM25=0.2, EMB=0.8). Single-shot rewriting matches max-10 iterative refinement within $0.2\%$.}
\label{fig:iteration_budget}
\end{figure}

\subsection{Optimization regimes diverge between retrieval-based and fixed-pool routing}
\label{subsec:regime}

Closed-world baseline F1 is high (\ensuremath{\sim}$91\%$ on I2) and gains from optimization are correspondingly modest ($0.6$-$0.9\%$ across I1, I2, I3, Appendix~\ref{sec:appendix-closedworld}).
Open-world baseline F1 is lower (\ensuremath{\sim}$70\%$ on I2) and gains are substantially larger ($+4.5\%$, Appendix~\ref{sec:appendix-openworld}).
More striking than the magnitude difference is a directional one: LLM-initialized descriptions \emph{hurt} closed-world F1 by $0.8$-$1.4\%$ comparing to the original placeholder text, while \emph{improving} open-world F1 by $1.5$-$2.1\%$.

The two regimes require separate optimization.
We re-evaluated closed-world-optimized descriptions in the open-world setting without further refinement:  gains are modest ($0.4$-$1.0\%$), well below in-setting open-world optimization ($3.7$-$4.5\%$).
Descriptions tuned against a fixed candidate pool do not generalize to the dynamic retrieved pool: closed-world optimization improves routing precision among a small known set of competitors, while open-world optimization additionally improves retrieval recall.

Description optimization is qualitatively different in retrieval-based vs.\ fixed-pool routing.
Practitioners should determine which regime their deployment matches and optimize within that regime: LLM initialization is a useful starting point for retrieval-based routing but should be avoided when the candidate pool is fixed and small.

\subsection{Initial training F1 as a diagnostic signal}
\label{subsec:overfit}

\begin{figure}[t]
\centering
\includegraphics[width=0.8\columnwidth]{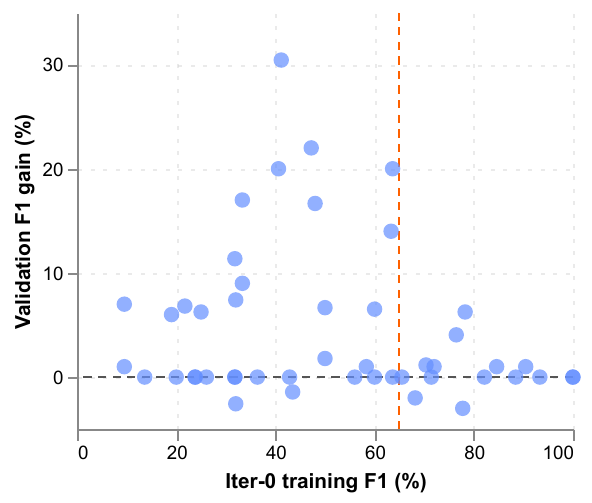}
\caption{Each dot is one ToolBench I2 tool ($n=48$). Tools with iter-0 training F1 below $65\%$ (vertical dashed line) achieve substantially higher validation F1 gains.}
\label{fig:overfit_scatter}
\end{figure}

Figure~\ref{fig:overfit_scatter} examines whether iter-0 training F1 (the routing F1 of the original placeholder description on the training set, before any optimization) predicts the value of optimization on held-out data.
On ToolBench open-world, tools with iter-0 training F1 below $65\%$ ($n=33$) achieve an average held-out gain of $+6.27\%$, approximately $10$ times the gain of tools with iter-0 training F1 at or above $65\%$ ($n=15$, $+0.63\%$; t-test $p < 0.001$).
The pattern is robust to threshold choice (Appendix~\ref{sec:appendix-overfit}), and all tools with validation gains above $10\%$ fall in the low iter-0 region.
Iter-0 training F1 is a useful prioritization signal: tools below $65\%$ are the candidates where optimization adds substantial value.
 
On ToolBench, most training improvements transfer to held-out performance.
Per-tool training and validation F1 gains on ToolBench are positively correlated (Spearman $\rho = 0.66$); only 2 of 48 tools ($4\%$) exhibit an overfitting signature where training gains substantially exceed validation gains (Appendix~\ref{sec:appendix-tradeoff}).
 
A subset of skills cannot be resolved by description optimization alone.
On the production agent, two skills (\texttt{IntKnowledge} and \texttt{WorkbackPlan}) follow this pattern: low iter-0 training F1, large training gains, but failure to exceed the HUMAN baseline on the held-out test (Table~\ref{tab:prod_human}; multi-seed confirmation in Appendix~\ref{sec:appendix-multiseed}).
Both skills carry inherent semantic overlap with other skills in the deployment: \texttt{IntKnowledge} retrieves person-related organizational data, overlapping \texttt{PeopleSearch}; \texttt{WorkbackPlan} produces structured plans, overlapping \texttt{StatusReport} in surface form.
Description text alone cannot disambiguate skills whose intended scopes overlap, regardless of training signal quality.
Practitioners encountering a skill with low iter-0 training F1 and a large $\Delta$train versus $\Delta$val gap should consider architectural intervention (e.g., intent-specific routing rules, restructured skill scopes) rather than description refinement.
 
\section{Conclusion}
\label{sec:conclusion}

We deployed an automated description optimization pipeline on a production enterprise group chat agent and validated findings at scale on ToolBench.
Systematic ablation showed most of the originally designed mechanisms are unnecessary; a single-pass LLM rewrite given any FP/FN cases captures the bulk of available improvement and is non-inferior to manually tuned descriptions at production scale.

\typeout{MAIN CONTENT ENDS ON PAGE \thepage}
\section*{Limitations}
\label{sec:limitations}

We acknowledge several limitations.
Our production system has 9 skills and 372 synthetic test cases, limiting statistical power for per-skill claims.
The ToolBench original descriptions are placeholder text, which inflates the absolute gain numbers compared to systems with already-curated descriptions; the system's value in such settings remains to be characterized.
Open-world ablations are conducted on I2 only; I1 and I3 open-world experiments are left for future work.
The semantic overfitting observation in production is based on 2 of 9 skills; while ToolBench data falsifies the original ``low iter-0 predicts failure'' hypothesis at scale, the production-specific pattern in low-data settings warrants validation on additional production deployments.

% Acknowledgments omitted in review version (anonymous submission).
% \section*{Acknowledgments}

% Bibliography entries for the entire Anthology, followed by custom entries
%\bibliography{anthology,custom}
% Custom bibliography entries only
\bibliography{custom}

\appendix

\section{Ablation Tables}
\label{sec:appendix-ablations}

We ablate on the error signal type, iterative refinement, dual editing, and training size in Tables~\ref{tab:errorsignal}, \ref{tab:errorsignal-teams}, \ref{tab:singleshot}, \ref{tab:dual-tb}, \ref{tab:dual-teams}, and \ref{tab:tb-trainsize-open}.

\begin{table}[h]
\centering
\small
\caption{Error signal type ablation on I2 open-world (48-tool intersection, BM25=0.2/EMB=0.8, train\_size=20, max10). All conditions within 0.$09\%$.}
\label{tab:errorsignal}
\begin{tabular}{lrrr}
\toprule
Feedback signal & Orig & Best & $\Delta_{0\to F}$ \\
\midrule
FP + FN + TP (full)  & 70.6 & 75.1 & $+4.5$\increase \\
FP-only (TP + FP)    & 70.8 & 75.3 & $+4.5$\increase \\
FN-only (TP + FN)    & 70.8 & 75.3 & $+4.5$\increase \\
FP + FN (no TP)      & 70.8 & 75.4 & $+4.6$\increase \\
\bottomrule
\end{tabular}
\end{table}

\begin{table}[h]
\centering
\small
\caption{Error signal type ablation on the production agent (9 skills, trainmax). All conditions within $0.69\%$, inside production API non-determinism noise. $\Delta$ = avg.\ New F1 $-$ avg.\ OLD F1 across all 9 skills.}
\label{tab:errorsignal-teams}
\begin{tabular}{lr}
\toprule
Feedback signal & Avg.\ $\Delta$ \\
\midrule
FP + FN + TP (full)  & $+3.65$\increase \\
FP-only (TP + FP)    & $+2.99$\increase \\
FN-only (TP + FN)    & $+3.47$\increase \\
FP + FN (no TP)      & $+2.96$\increase \\
\bottomrule
\end{tabular}
\end{table}

\begin{table}[h]
\centering
\small
\caption{Single-shot vs.\ iterative refinement on I2 open-world (47-tool overlap, BM25=0.2/EMB=0.8, train\_size=20).}
\label{tab:singleshot}
\begin{tabular}{lrrr}
\toprule
Method & $n$ & Orig & $\Delta_{0\to F}$ \\
\midrule
Single-shot                     & 47 & 71.0 & $+4.4$\increase \\
Iterative, sampled (max10)      & 47 & 70.7 & $+4.6$\increase \\
Iterative, full-context (max10) & 47 & 70.9 & $+4.9$\increase \\
\bottomrule
\end{tabular}
\end{table}

\begin{table}[h]
\centering
\small
\caption{Dual editing ablation on ToolBench I2 open-world (BM25=0.2/EMB=0.8, train\_size=20, max10). $n$ differs because dual editing fails on one tool whose training cases trigger the Azure content filter. Difference of $0.03\%$ is well within experimental noise.}
\label{tab:dual-tb}
\begin{tabular}{lrrrr}
\toprule
Setting & $n$ & Orig & Best & $\Delta_{0\to F}$ \\
\midrule
Standard      & 48 & 70.6 & 75.1 & $+4.51$\increase \\
Dual editing  & 47 & 70.3 & 74.9 & $+4.54$\increase \\
\bottomrule
\end{tabular}
\end{table}

\begin{table}[h]
\centering
\small
\caption{Dual editing ablation on the production agent (trainmax). $\Delta$ = Iter F1 $-$ Avg OLD baseline (averaged across 4 settings, $75.6\%$ overall). Average effect of dual editing is $-0.11\%$, well within the per-skill $\pm 0.78\%$ noise floor.}
\label{tab:dual-teams}
\begin{tabular}{lrr}
\toprule
Skill & Std $\Delta$ & Dual $\Delta$ \\
\midrule
IntKnowledge   & $-1.7$\decrease  & $-3.5$\decrease  \\
PeopleSearch   & $+13.3$\increase & $+12.5$\increase \\
Email          & $+1.0$\increase  & $-0.3$\decrease         \\
UserManager    & $+7.5$\increase  & $+6.9$\increase  \\
WorkbackPlan   & $-0.1$\decrease           & $+0.6$\increase           \\
DirectReports  & $+3.2$\increase  & $+3.8$\increase  \\
WebSearch      & $+5.7$\increase  & $+5.8$\increase  \\
Calendar       & $+2.7$\increase  & $+3.6$\increase  \\
StatusReport   & $+0.9$           & $+2.0$\increase  \\
\midrule
\textbf{Average} & $\mathbf{+3.6}$\increase & $\mathbf{+3.5}$\increase \\
\bottomrule
\end{tabular}
\end{table}

\begin{table}[h]
\centering
\small
\caption{Training size ablation on ToolBench I2 open-world (BM25=0.2/EMB=0.8, max10). $n$ varies because tool eligibility (at least one retrieval false positive in the train set) increases with train size. $\Delta$ values within $0.5\%$.}
\label{tab:tb-trainsize-open}
\begin{tabular}{lrrrr}
\toprule
Train size & $n$ & Orig & Best & $\Delta_{0\to F}$ \\
\midrule
5  & 31 & 67.8 & 72.7 & $+4.9$\increase \\
10 & 42 & 69.8 & 74.8 & $+5.0$\increase \\
20 & 48 & 70.6 & 75.1 & $+4.5$\increase \\
\bottomrule
\end{tabular}
\end{table}

\section{Open-World Detail (ToolBench)}
\label{sec:appendix-openworld}

Table~\ref{tab:open} reports the full open-world results referenced in \S\ref{subsec:regime}.
\textbf{Transfer} re-evaluates closed-world-optimized descriptions in the open-world retrieval setting without further refinement.
\textbf{In-setting} runs the full optimization loop using the retrieved candidate pool.
Embedding-heavy retrieval (BM25=0.2, EMB=0.8) consistently outperforms BM25-heavy across both transfer and in-setting; BM25 fails to differentiate tools with generic placeholder names.
$n$ is the tools with at least one retrieval false positive; transfer evaluates all 100 tools (no FP requirement), while in-setting requires at least one FP per tool to be optimizable.

\begin{table}[h]
\centering
\small
\caption{Open-world results on ToolBench I2. Transfer re-evaluates closed-world descriptions; in-setting re-optimizes against the retrieved candidate pool.}
\label{tab:open}
\begin{tabular}{llrrrr}
\toprule
Setting & BM25/EMB & $n$ & Orig & Best & $\Delta_{0\to F}$ \\
\midrule
\multicolumn{6}{l}{\textit{Transfer (re-evaluate closed-world descriptions)}} \\
\midrule
 & 0.2/0.8 & 100 & 70.9 & 71.8 & $+1.0$ \\
 & 0.5/0.5 & 100 & 71.8 & 72.6 & $+0.7$ \\
 & 0.8/0.2 & 100 & 70.0 & 70.5 & $+0.4$ \\
\midrule
\multicolumn{6}{l}{\textit{In-setting optimization (max10)}} \\
\midrule
 & 0.2/0.8 & 48 & 70.6 & \textbf{75.1} & $\mathbf{+4.5}$\increase \\
 & 0.5/0.5 & 45 & 68.9 & 72.8 & $+3.9$\increase \\
 & 0.8/0.2 & 43 & 70.0 & 73.7 & $+3.7$\increase \\
\bottomrule
\end{tabular}
\end{table}

\section{Production train20 Results}
\label{sec:appendix-train20}

Trainmax (used in the main paper, \S\ref{subsec:noninferior}) uses all available positive examples per skill (10 to 119, varying by skill). For comparison, train20 caps the training set at 20 positive plus 20 negative examples per skill (or fewer when the skill has fewer positives available; \texttt{SendEmail}, \texttt{generate\_status\_report}, \texttt{WorkbackPlan} fall back to all available data).
Table~\ref{tab:prod_train20} reports per-skill F1 for the train20-standard run.
On average, train20 yields $\Delta$(OLD$\to$Iter) $= +1.71\%$ versus trainmax-standard's $+3.65\%$, indicating that more training data helps in aggregate but not uniformly: 5 skills are within $1\%$ of their trainmax result, while \texttt{GetUserManager} regresses sharply at train20 (an outlier we attribute to API non-determinism, since the multi-seed trainmax results show \texttt{GetUserManager} is otherwise stable; Appendix~\ref{sec:appendix-multiseed}).

\begin{table}[h]
\centering
\small
\caption{train20-standard production results (per-skill F1; train pos/neg = 20/20, capped by availability). $\Delta$ = Iter $-$ OLD.}
\label{tab:prod_train20}
\begin{tabular}{lrrrr}
\toprule
Skill & OLD & INIT & Iter & $\Delta$ \\
\midrule
PeopleSearch    & 66.7 & 78.4 & 78.5 & $+11.8$\increase \\
WebSearch       & 73.0 & 78.5 & 78.0 & $+5.0$\increase  \\
Calendar        & 75.2 & 75.7 & 77.0 & $+1.8$\increase  \\
IntKnowledge    & 82.6 & 78.7 & 79.1 & $-3.4$\decrease  \\
UserManager     & 71.5 & 78.9 & 66.7 & $-4.9$\decrease  \\
DirectReports   & 76.0 & 79.3 & 79.3 & $+3.2$\increase  \\
Email           & 79.1 & 78.8 & 79.6 & $+0.6$\increase           \\
StatusReport    & 77.1 & 77.8 & 78.1 & $+1.0$\increase           \\
WorkbackPlan    & 80.4 & 79.0 & 79.3 & $-1.1$\decrease  \\
\midrule
\textbf{Average} & & & & $\mathbf{+1.7}$\increase \\
\bottomrule
\end{tabular}
\end{table}

\section{OLD Baseline (Production Agent)}
\label{sec:appendix-oldbaseline}

OLD measures the system-level F1 with the new skill disabled, evaluated on the subset of test cases that do not target the skill.
Because OLD uses a different test subset than HUMAN/INIT/Iter (which evaluate on the full 372 cases with the skill enabled), OLD$\to$Iter gains conflate skill availability with description quality and are not directly comparable to HUMAN$\to$Iter.
We report OLD here for completeness; \S\ref{subsec:noninferior} uses HUMAN$\to$Iter as the primary comparison.

\begin{table*}[h]
\centering
\small
\caption{Production results (trainmax, standard optimization) including the OLD baseline. OLD evaluates the system with the new skill disabled on a different test subset; it is not directly comparable to HUMAN/INIT/NEW. $\Delta$Deploy = OLD$\to$NEW (deployment + quality). Averaged OLD baseline across runs.}
\label{tab:prod_old}
\begin{tabular}{lrrrrr}
\toprule
Skill & OLD & HUMAN & INIT & NEW & $\Delta$Deploy \\
\midrule
PeopleSearch   & 66.7 & 79.4 & 79.2 & 79.9 & $+13.3$\increase \\
WebSearch      & 72.4 & 79.4 & 78.4 & 78.1 & $+5.7$\increase  \\
Calendar       & 75.3 & 79.4 & 77.6 & 77.9 & $+2.7$\increase  \\
IntKnowledge   & 82.7 & 79.4 & 77.1 & 81.0 & $-1.7$\decrease  \\
UserManager    & 72.2 & 79.4 & 79.4 & 79.7 & $+7.5$\increase  \\
DirectReports  & 75.4 & 79.4 & 79.5 & 78.6 & $+3.2$\increase  \\
Email          & 78.8 & 79.4 & 78.6 & 79.8 & $+1.0$\increase  \\
StatusReport   & 76.8 & 79.4 & 77.6 & 77.7 & $+0.9$\increase  \\
WorkbackPlan   & 79.8 & 79.4 & 78.3 & 79.7 & $-0.1$           \\
\midrule
\textbf{Average} & & \textbf{79.4} & 78.4 & 79.2 & $\mathbf{+3.6}$\increase \\
\bottomrule
\end{tabular}
\end{table*}

\section{Multi-Seed Variance (Production Agent)}
\label{sec:appendix-multiseed}

To characterize run-to-run variance, we ran 3 independent seeds of the trainmax-standard pipeline.
Each seed includes a fresh optimization run (LLM calls at temperature $> 0$) and a fresh regression evaluation (live API calls at temperature $> 0$).
The HUMAN baseline ($79.4\%$) was evaluated in a single regression run; we did not characterize HUMAN-specific run variance.
The reported per-skill noise floor of $\Delta$F1 ($\pm 0.78\%$ std on average; Table~\ref{tab:multiseed}) bounds Iter variance directly; $\Delta$(HUMAN$\to$Iter) variance includes additional HUMAN evaluation noise we did not measure.
The mean $\Delta$(OLD$\to$Iter) is $+3.27 \pm 0.78\%$ across the 3 seeds, with 8 of 9 skills below $\pm 1\%$ std.
\texttt{IntKnowledge} regresses against OLD in all 3 seeds ($-1.90\%$, $-3.72\%$, $-2.04\%$), confirming this is a robust failure rather than a single-run artifact.

\begin{table*}[h]
\centering
\small
\caption{Per-skill $\Delta$F1 (NEW $-$ OLD) across 3 independent seeds of the trainmax-standard pipeline. Std is sample standard deviation across seeds. The average per-skill std is $\pm 0.78\%$.}
\label{tab:multiseed}
\begin{tabular}{lrrrrr}
\toprule
Skill & Seed 1 & Seed 2 & Seed 3 & Mean & Std \\
\midrule
PeopleSearch  & $+13.0$ & $+10.7$ & $+11.8$ & $+11.8$ & $\pm 1.2$ \\
WebSearch     & $+5.7$  & $+6.2$  & $+5.7$  & $+5.9$  & $\pm 0.3$ \\
Calendar      & $+2.7$  & $+1.9$  & $+2.8$  & $+2.5$  & $\pm 0.5$ \\
IntKnowledge  & $-1.9$  & $-3.7$  & $-2.0$  & $-2.6$  & $\pm 1.0$ \\
UserManager   & $+6.5$  & $+7.1$  & $+5.5$  & $+6.4$  & $\pm 0.8$ \\
DirectReports & $+4.2$  & $+3.8$  & $+2.8$  & $+3.6$  & $\pm 0.8$ \\
Email         & $+0.6$  & $+1.3$  & $-0.2$  & $+0.6$  & $\pm 0.8$ \\
StatusReport  & $+1.9$  & $+0.9$  & $+0.1$  & $+1.0$  & $\pm 0.9$ \\
WorkbackPlan  & $+0.1$  & $+1.2$  & $-0.4$  & $+0.3$  & $\pm 0.8$ \\
\midrule
\textbf{Average} & $\mathbf{+3.6}$ & $\mathbf{+3.3}$ & $\mathbf{+2.9}$ & $\mathbf{+3.3}$ & $\mathbf{\pm 0.8}$ \\
\bottomrule
\end{tabular}
\end{table*}

\section{Production Training F1 Dynamics}
\label{sec:appendix-trainf1}

Table~\ref{tab:teams-trainf1} reports per-skill training-set F1 at iteration 0 (the initial LLM-generated description) and the best across iterations on the trainmax-standard production setting.
Two skills exhibit large training F1 gains: \texttt{IntKnowledge} ($+30.8$pp) and \texttt{WorkbackPlan} ($+28.1$pp).
Both also start with low iter-0 training F1 ($\leq 67\%$), and neither translates the training gain into a held-out F1 improvement above the HUMAN baseline (Table~\ref{tab:prod_human}).
The other 7 skills show no training F1 movement: each begins at or above the $90\%$ stop criterion and the optimization loop terminates at iter-0.
The pattern matches the small-data overfitting characterization in \S\ref{subsec:overfit}.

\begin{table}[h]
\centering
\small
\caption{Per-skill training-set F1 at iter-0 (initial LLM description) vs.\ best across iterations on the production agent (trainmax-standard). $\Delta$train is the iterative refinement gain on the training subset. Skills ordered as in Table~\ref{tab:prod_human}.}
\label{tab:teams-trainf1}
\begin{tabular}{lrrr}
\toprule
Skill & iter-0 & best & $\Delta$train \\
\midrule
IntKnowledge   & 58.1  & 88.9  & $\mathbf{+30.8}$\increase \\
PeopleSearch   & 97.0  & 97.0  & $+0.0$ \\
Email          & 90.3  & 90.3  & $+0.0$ \\
UserManager    & 96.2  & 96.2  & $+0.0$ \\
WorkbackPlan   & 66.7  & 94.7  & $\mathbf{+28.1}$\increase \\
DirectReports  & 95.5  & 95.5  & $+0.0$ \\
WebSearch      & 92.6  & 92.6  & $+0.0$ \\
Calendar       & 90.0  & 90.0  & $+0.0$ \\
StatusReport   & 100.0 & 100.0 & $+0.0$ \\
\bottomrule
\end{tabular}
\end{table}

\section{Iter-0 Training F1 as a Failure Predictor (ToolBench)}
\label{sec:appendix-overfit}

We test whether the production observation in \S\ref{subsec:overfit}, that low iter-0 training F1 predicts failure, generalizes to the 48 ToolBench open-world tools (I2, BM25=0.2/EMB=0.8, train\_size=20, max10).
For each tool, we extract iter-0 training F1 (\textbf{train\_f1@0}) and the held-out validation gain $\Delta$val from the existing optimization logs; no new LLM calls are required.

The 65\% threshold derived from production is highly significant on ToolBench (Welch's $t$, $p = 0.0008$, $N=48$): tools below 65\% iter-0 training F1 gain $+6.3\%$ on validation on average, vs.\ $+0.6\%$ for tools above (Table~\ref{tab:overfit}).
However, the direction is opposite to the production failure mode: at ToolBench scale, low iter-0 F1 predicts \emph{larger} held-out gains, not failure to generalize.
Spearman correlation between $\Delta$train and $\Delta$val is $+0.66$ ($p < 10^{-4}$): in the open-world retrieval setting, large training gains generalize.
We interpret the production overfitting (\texttt{IntKnowledge}, \texttt{WorkbackPlan}) as a small-data-regime artifact of low-positive-sample skills (10--119 positives), not a general property of the pipeline.

\begin{table*}[h]
\centering
\small
\caption{Threshold sweep: mean $\Delta$val for tools below vs.\ above each iter-0 training F1 threshold $\tau$. The 65\% threshold from production is highly significant ($p = 0.0008$). At ToolBench scale, low iter-0 F1 predicts \emph{larger} validation gains, not failure.}
\label{tab:overfit}
\begin{tabular}{rrrrrr}
\toprule
$\tau$ & $n_{<\tau}$ & mean $\Delta$val$_{<\tau}$ & $n_{\geq\tau}$ & mean $\Delta$val$_{\geq\tau}$ & $p$ \\
\midrule
55\% & 26 & $+6.4$ & 22 & $+2.3$ & $0.052$ \\
60\% & 28 & $+5.9$ & 20 & $+2.5$ & $0.093$ \\
\textbf{65\%} & \textbf{33} & $\mathbf{+6.3}$ & \textbf{15} & $\mathbf{+0.6}$ & $\mathbf{0.0008}$ \\
70\% & 35 & $+5.9$ & 13 & $+0.9$ & $0.002$ \\
75\% & 38 & $+5.5$ & 10 & $+0.9$ & $0.005$ \\
80\% & 41 & $+5.2$ & 7  & $+0.3$ & $0.0003$ \\
\bottomrule
\end{tabular}
\end{table*}

\section{Closed-World Results}
\label{sec:appendix-closedworld}

Table~\ref{tab:closed} shows validation F1 in the closed-world setting, where the candidate pool is the per-query \texttt{available\_tools} list (3--15 tools) rather than the full retrieved corpus.
A consistent pattern emerges across all splits: LLM initialization (\textbf{init}) \emph{reduces} F1 by $0.8$--$1.4\%$ relative to the original placeholder descriptions (\textbf{orig}), while iterative refinement recovers and surpasses the original.
Despite being completely uninformative text, placeholder descriptions perform comparably to zero-shot LLM-generated ones in this setting; the gains come entirely from error feedback, not initialization.
This contrasts with the open-world setting (\S\ref{subsec:regime}) where initialization helps, suggesting description quality matters more when retrieval recall is the bottleneck.

On I2, the training size sweet spot is 10 examples per class ($+1.0\%$); additional examples do not consistently improve the refinement signal.
Iterative refinement ($+0.9\%$) modestly outperforms single-shot ($+0.6\%$), consistent with per-example signal being weaker in the fixed small candidate pool.

\begin{table*}[h]
\centering
\small
\caption{Closed-world results. Top: I1/I2/I3 under standard optimization (train\_size=20, max10). Bottom: I2 training size ablation and single-shot comparison. Val F1 averaged over 100 tools.}
\label{tab:closed}
\begin{tabular}{lrrrrrr}
\toprule
 & $n$ & Orig & Init & Best & $\Delta_{0\to1}$ & $\Delta_{0\to F}$ \\
\midrule
I1 & 100 & 90.6 & 89.4 & 91.5 & $-1.2$ & $+0.9$\increase \\
I2 & 100 & 90.6 & 89.2 & 91.5 & $-1.4$ & $+0.9$\increase \\
I3 & 100 & 82.8 & 82.0 & 83.5 & $-0.8$ & $+0.6$\increase \\
\midrule
\multicolumn{7}{l}{\textit{I2 training size ablation}} \\
\midrule
train=5  & 100 & 90.7 & 89.5 & 91.1 & $-1.2$ & $+0.4$\increase \\
train=10 & 100 & 90.7 & 89.3 & \textbf{91.7} & $-1.4$ & $\mathbf{+1.0}$\increase \\
train=20 & 100 & 90.6 & 89.2 & 91.5 & $-1.4$ & $+0.9$\increase \\
\midrule
\multicolumn{7}{l}{\textit{I2 single-shot vs.\ iterative (train=20, max10)}} \\
\midrule
Single-shot       & 100 & 90.6 & —    & 91.3 & —      & $+0.6$\increase \\
Iterative         & 100 & 90.6 & 89.2 & 91.5 & $-1.4$ & $+0.9$\increase \\
\bottomrule
\end{tabular}
\end{table*}

\section{Training versus validation gain at ToolBench scale}
\label{sec:appendix-tradeoff}
 
\begin{figure}[t]
\centering
\includegraphics[width=0.8\columnwidth]{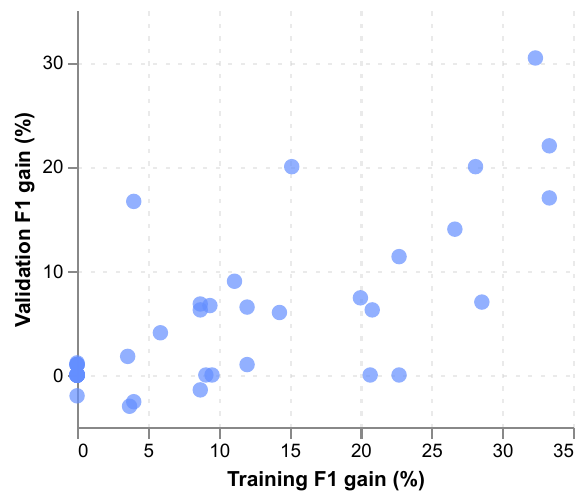}
\caption{Per-tool $\Delta$train versus $\Delta$val on ToolBench I2 open-world ($n=48$). Spearman $\rho = 0.66$. Two outliers in the lower-right exhibit overfitting.}
\label{fig:toolbench_train_vs_val}
\end{figure}

Figure~\ref{fig:toolbench_train_vs_val} plots per-tool training F1 gain against validation F1 gain for the 48 tools optimized in the open-world setting.
Training and validation gains are positively correlated (Spearman $\rho = 0.66$); 2 tools ($4\%$) exhibit an overfitting signature ($\Delta$train $> 20\%$, $\Delta$val $< 5\%$).
 
\end{document}